\documentclass[]{oppo}

\usepackage[toc,page,header]{appendix}

\usepackage{minitoc}

\usepackage{graphicx}

\usepackage{amsmath,booktabs, multirow,makecell,multicol,xspace}

\newcommand{\figref}[1]{Figure \ref{#1}}
\newcommand{\tabref}[1]{Table \ref{#1}}

\newcommand{\secref}[1]{Section \ref{#1}}

\tcbset{
  aibox/.style={
    width=\linewidth,
    top=8pt,
    bottom=4pt,
    colback=green!3!white,
    colframe=black,
    colbacktitle=black,
    enhanced,
    center,
    attach boxed title to top left={yshift=-0.1in,xshift=0.15in},
    boxed title style={boxrule=0pt,colframe=white,},
  }
}
\newtcolorbox{AIbox}[2][]{aibox,title=#2,#1}

\title{\includegraphics[width=1.2cm]{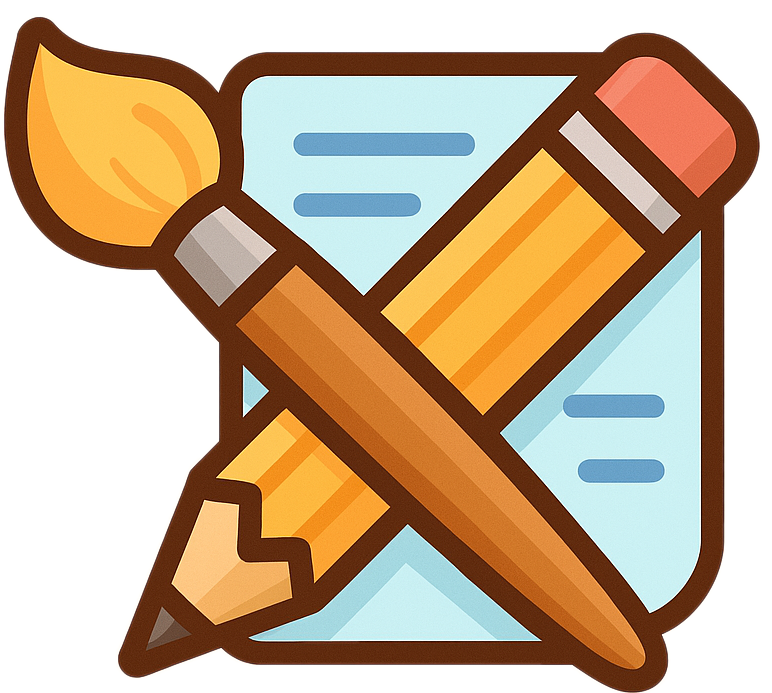} \textsc{TaskCraft}: Automated Generation of Agentic Tasks} 

\affiliation{OPPO AI Agent Team}

\contribution[]{Full author list in Contributions}
\abstract{
Agentic tasks, which require multi-step problem solving with autonomy, tool use, and adaptive reasoning, are becoming increasingly central to the advancement of NLP and AI. However, existing instruction data lacks tool interaction, and current agentic benchmarks rely on costly human annotation, limiting their scalability. We introduce \textsc{TaskCraft}, an automated workflow for generating difficulty-scalable, multi-tool, and verifiable agentic tasks with execution trajectories. TaskCraft expands atomic tasks using depth-based and width-based extensions to create structurally and hierarchically complex challenges. Empirical results show that these tasks improve prompt optimization in the generation workflow and enhance supervised fine-tuning of agentic foundation models. We present a large-scale synthetic dataset of approximately 36,000 tasks with varying difficulty to support future research on agent tuning and evaluation.
}

\date{\today}
\correspondence{\email{zhouwangchunshu@oppo.com}}

\checkdata[Code \& Data]{\url{https://github.com/OPPO-PersonalAI/TaskCraft}}

\begin{document}
\maketitle


\section{Introduction}

Agentic tasks—autonomous, multi-step problem-solving requiring tool use and adaptive reasoning—are increasingly pivotal in AI and NLP. Advances in language agents~\cite{autogpt2023autogpt,wu2023autogen,li2023camel,zhou2023recurrentgpt,zhou2023agents,zhou2024agents2} have shifted AI from passive assistance to proactive agency, enabling complex workflow execution. This is exemplified by systems combining reasoning frameworks like ReAct~\cite{yao2023react} with dynamic orchestration, where solution trajectories critically improve inference quality. However, the inherent complexity of such tasks challenges conventional annotation paradigms, necessitating novel approaches to model training and evaluation.

To assess advanced agent capabilities, benchmarks such as GAIA~\cite{mialon2023gaia}, BrowseComp~\cite{wei2025browsecomp}, and Humanity’s Last Exam (HLE)~\cite{phan2025humanitysexam} have been introduced. GAIA evaluates reasoning, tool use, and web browsing through 466 real-world questions. BrowseComp comprises 1,266 tasks that test an agent’s ability to retrieve and integrate complex online information. HLE includes 2,500 multi-modal questions across over 100 disciplines to measure advanced reasoning and domain knowledge. While these datasets have significantly contributed to agent evaluation, they suffer from scalability limitations due to the labor-intensive nature of data annotation. For example, creating HLE required 1,000 experts to label just 2,500 data points, hindering its ability to scale.

Prior work has explored the automatic generation of instruction-following data using large language models to alleviate the scalability issues of human-annotated datasets. A representative example is the Self-Instruct framework~\cite{wang2022self}, which demonstrated that LLMs can generate high-quality, diverse instruction data for multi-turn dialogues. This approach has proven effective for supervised fine-tuning (SFT). However, these methods are primarily designed for static instruction-following scenarios and fall short in modeling agentic tasks, which require interaction with external tools and environments. Consequently, such data is insufficient for training or evaluating agents that operate in dynamic, real-world settings.

In this work, we introduce \textsc{TaskCraft}\xspace, an agentic workflow for the automated generation of agentic tasks. Our approach provides the following advantages:

\begin{itemize}
    \item \textbf{Scalability.} The workflow supports adaptive difficulty, seamless multi-tool integration, and the generation of tasks beyond the capabilities of the task-generation agent, along with their corresponding trajectories.
    \item \textbf{Efficient Verification.} During each task extension, only incremental components undergo agentic validation, eliminating the need for full verification of the extended task.
\end{itemize}

The core approach involves initially generating multiple atomic tasks, each solvable with a single target tool invocation, and then expanding them using depth-based and width-based extension. For depth-based task extension, we iteratively transform specific textual elements of the original task (such as key terms) into a new atomic task to support progressive resolution. In contrast, the width-based extension formulates tasks that require resolving multiple sub-problems by integrating distinct problem instances. 

To ensure high-quality agentic tasks, we employ a rejection sampling strategy during verification. For atomic tasks, we include cases where an agent using external tools can solve the task while an LLM cannot, ensuring that atomic tasks genuinely necessitate tool usage.
For extension tasks, we leverage linguistic analysis with LLMs, enabling rapid validation and facilitating the creation of challenges beyond existing agent capabilities. This approach enhances efficiency and broadens problem-solving potential.

The controlled generation process ensures inherent access to ground-truth execution trajectories, enabling precise interpretability, reproducibility, and verifiability—critical for agent evaluation and reinforcement learning. To further validate task effectiveness, we implement a self-evolving prompt optimization strategy inspired by bootstrap few-shot learning~\cite{khattab2024dspy}. This iterative refinement improves rejection sampling pass rates while minimizing generation time. Additionally, we leverage the generated task trajectories to train an agent foundation model~\cite{search-r1}. Experimental results show that an independent LLM, trained on these trajectories, effectively plans and invokes tools, yielding performance gains on HotpotQA~\cite{yang2018hotpotqa}, Musique~\cite{trivedi2022musique}, and Bamboogle~\cite{press2022measuring}.

Based on this method, we generated a task dataset comprising approximately 36,000 tasks of varying difficulty, each requiring different tools for resolution, including search, web browsing, PDF reading, and image understanding.

Our key contributions are as follows: 
\begin{itemize}
    \item We introduce an automated agentic task generation workflow capable of producing 
    scalable difficulty, efficient verification, and multi-tool supported tasks, along with their corresponding execution trajectories.
    \item We empirically evaluate task effectiveness through prompt learning, which facilitates the self-evolution of our workflow and holds potential for optimizing existing agent workflows. Additionally, SFT is applied to an agent foundation model, enabling it to substitute agent workflows where applicable.
    \item We release a synthetic dataset comprising about 32k agentic tasks of varying difficulty levels, complete with their execution trajectories, to facilitate further research.
\end{itemize}

\section{Notations and Preliminary}

\begin{figure}[htbp]
   \centering
   \includegraphics[width=0.43\textwidth]{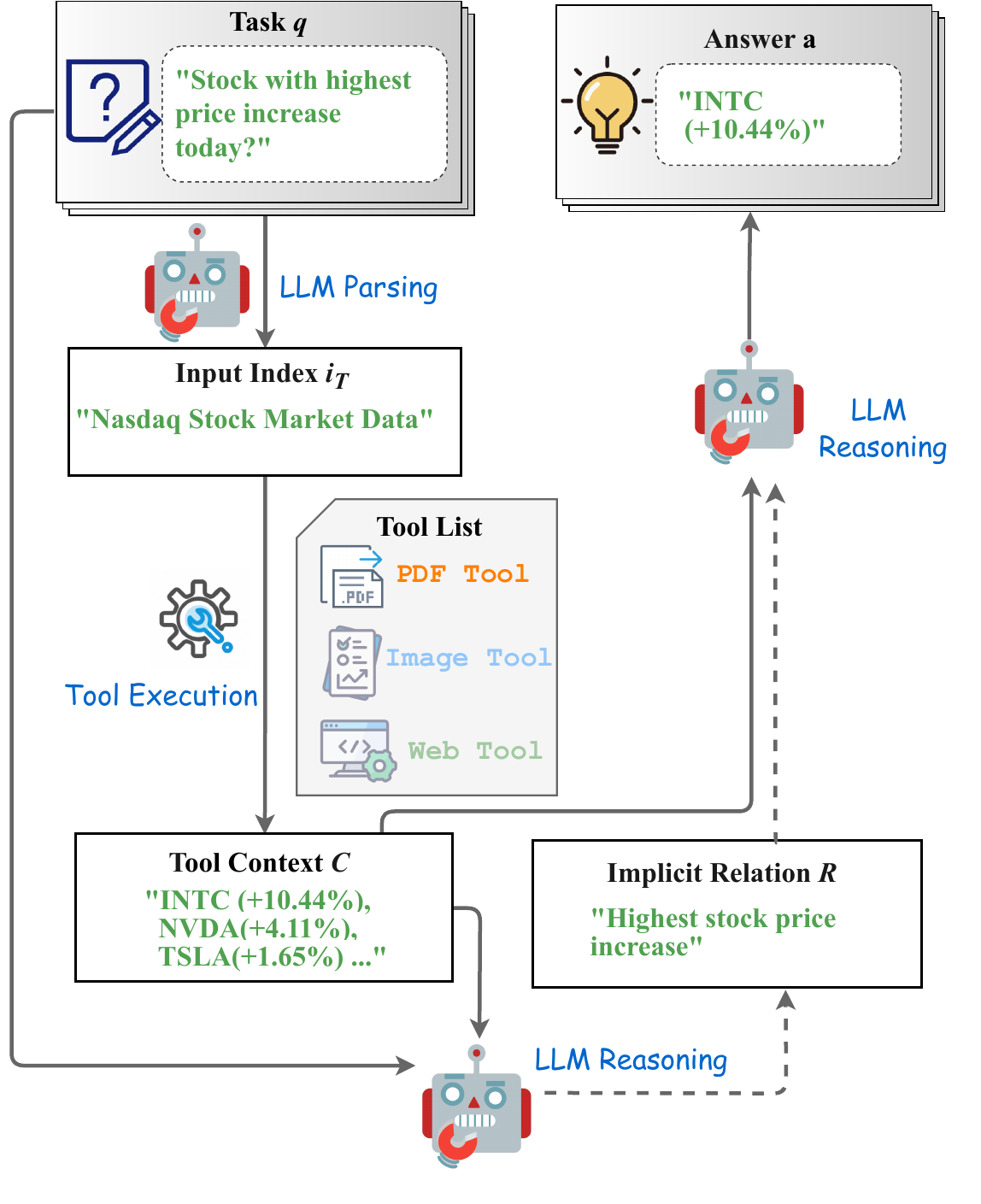}
   \caption{Execution flow of a single tool invocation. The agent extracts the input index \( i_T \) (e.g., document name, webpage title) for invoking tool \( T \), focusing solely on steps that yield valid tool context. Executing \( T \) with \( i_T \) retrieves context \( C \), enabling the LLM to infer the relationship \( R \) and produce the final result \( a \).}  
   \label{fig:tool_exec1}          
\end{figure}

\begin{AIbox}
    {\makecell{Tool-Assisted Task Execution}}{As \figref{fig:tool_exec1} shown, given a task \( q \), the agent extracts the input index \( i_T \) (e.g., document name, webpage title) for invoking a target tool \( T \). We focus solely on steps that yield a valid tool context, omitting unrelated processes such as file location or search for simplicity. Executing tool \( T \) with \( i_T \) retrieves the associated context \( C \). The LLM implicitly deduces the relationship \( R \) between \( C \) and the expected outcome, producing the final result \( a \).}
\end{AIbox}



\begin{AIbox}
    {\makecell{Atomic Task}}{An atomic task is resolved with a single target tool invocation. To simplify, we disregard search and file system operations, assuming a detailed input index $i_T$ enables retrieval through finite navigation.}
\end{AIbox}

Given an answer $a$, the most direct approach to construct an atomic task involves prompting an LLM to generate the corresponding question. However, questions produced in this manner often suffer from low tool invocation rates, unpredictable difficulty levels, unregulated tool requirements, and inconsistent verification complexity (see \secref{sec:abla} for more details). 

To mitigate these issues, we assume an ideal search engine capable of retrieving precise data based on $i_T$ (e.g., paper titles, image paths, music names, etc.). Under this assumption, we can construct a task question $q= f(i_T, R) \xrightarrow{} a$, where $f$ represents a sampling function that enables the LLM to generate the corresponding natural language representation of the question $q$ based on the provided information.

\begin{figure*}[htbp]
   \centering
    \includegraphics[width=0.9\textwidth]{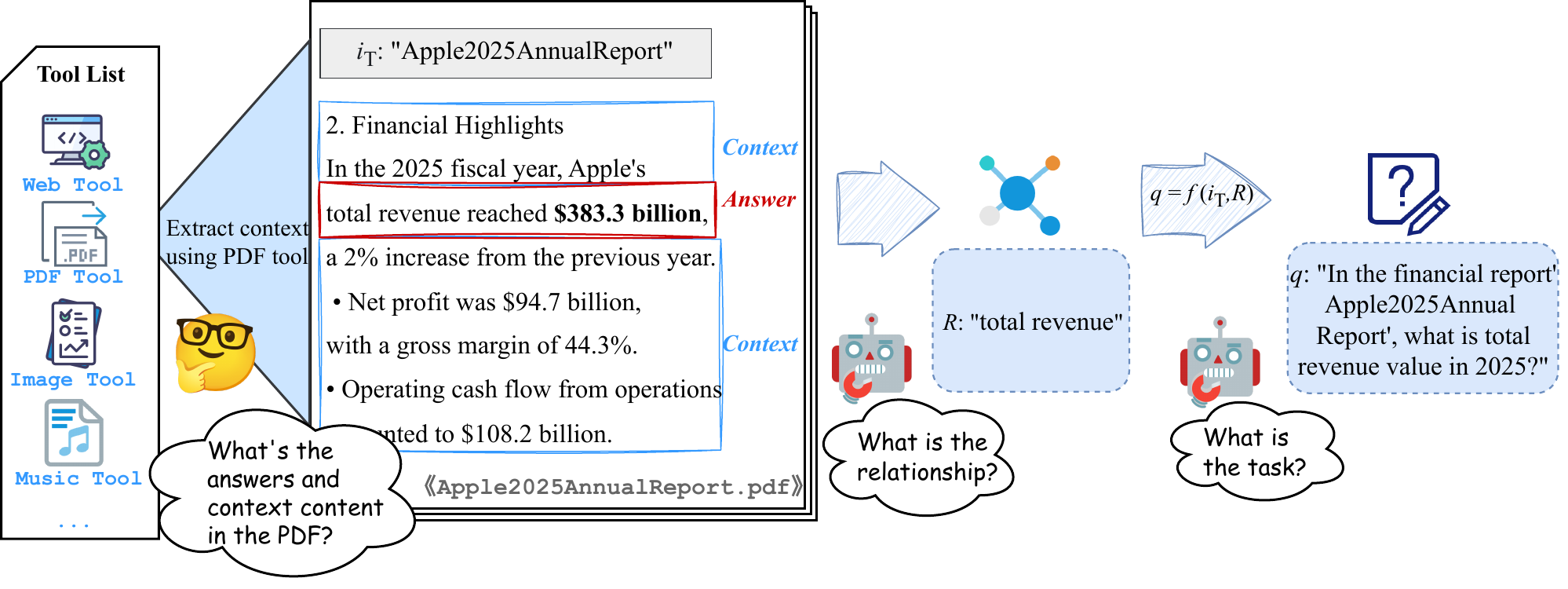}
    \caption{Atomic task generation: From an unlabeled corpus, we extract $i_T$ and derive textual content $C$ via tool execution. LLM identifies candidate answers $a$ from $C$, infers their relationship $R$, and constructs question $q$ conditioned on $i_T$ and $R$.}  
   \label{fig:aotomic_task}  
\end{figure*}

\section{Automated Task Generation Workflow}

\subsection{Atomic Task Generation}\label{sec:atomic}

As \figref{fig:aotomic_task} shown, we begin by compiling a corpus of unlabeled data aligned with the tool's input requirements. From this corpus, we extract $i_T$ and derive textual content $C$ via tool execution. For example, browsing, PDF, and image comprehension tools yield webpage titles, PDF names, and image paths, from which we extract textual content $C$ for answer sampling. We prompt an LLM to identify key candidate answers $a$ from $C$ and infer their relationship $R$ with $C$, ultimately constructing question $q$ conditioned on $i_T$ and $R$.

\subsection{Task Extension}\label{sec:extension}
In order to increase task difficulty in a scalable way, we adopted two extended task strategies: the \emph{depth-based extension} and the \emph{width-based extension}.


\noindent\textbf{Depth-based extension.} We aim to construct tasks requiring multiple sequential tool executions, where each step depends on the output of the previous one. To achieve this, a new subproblem must be derived from a known problem $q^n$. The tool input index $i_T$ at each stage exhibits strong extensibility due to (1) its frequent association with proper nouns, which are less likely to be memorized by LLMs, and (2) its natural suitability for recursive definition. Specifically, a single atomic task follows the formulation:
\begin{equation}
    q^n = f(i_T^n, R^n) \xrightarrow{} a.
\end{equation}

\begin{figure*}[htbp]
    \centering
    \includegraphics[width=\textwidth]{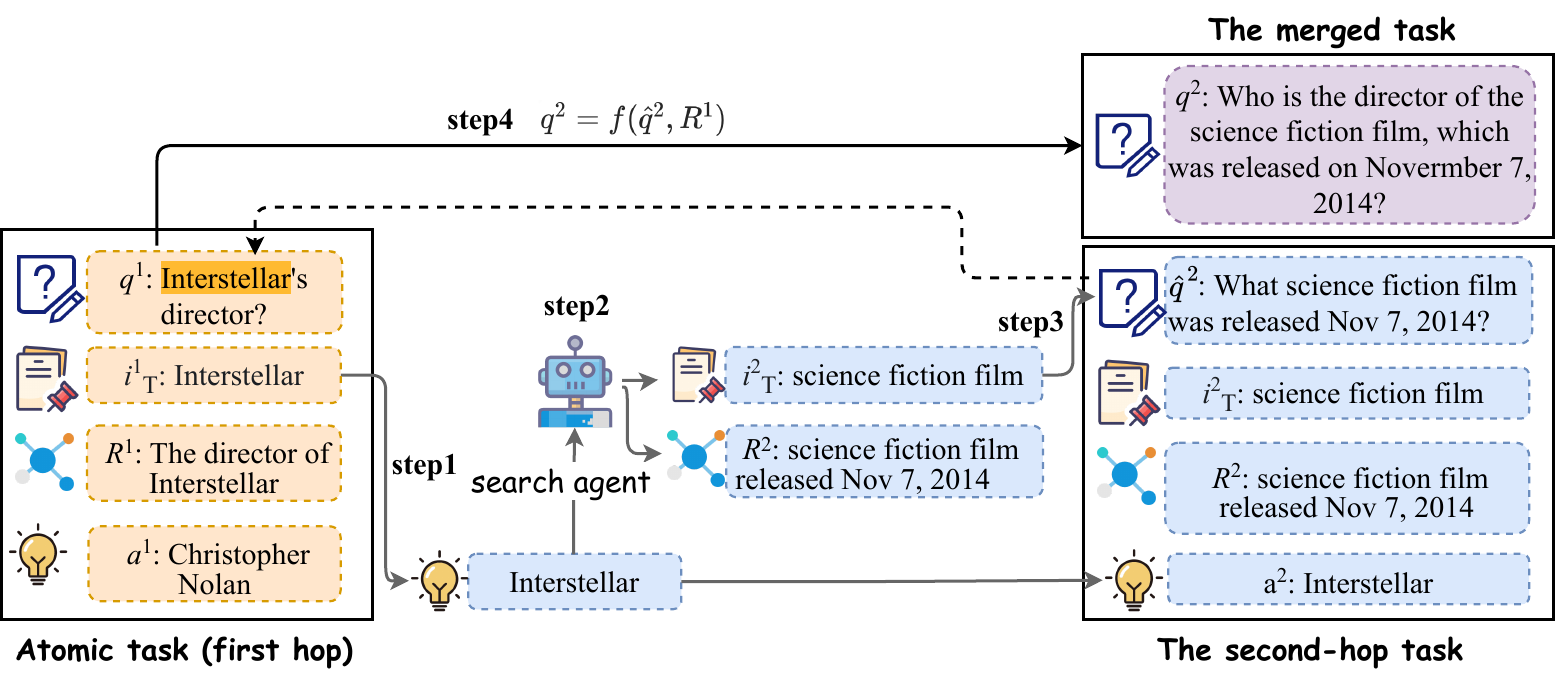}
   \caption{Depth-based extension. A 1-hop task $q^1$ is recursively extended to a 2-hop task $q^2$. A search agent derives the new tool input index $i_T^2$ by extracting superset candidates $C^2$, which an LLM analyzes to determine $i_T^2$ and its relationship $R^2$ with $i_T^1$. After verification, the refined question $q^2$ integrates $\hat{q}^2$ with historical relationships $R^1$.}
   \label{fig:depth-merged}
\end{figure*}

To extend a n-hot task $q^n$ into a (n+1)-hop dependency task $q^{n+1}$, we can define the recursive formulation:
\begin{equation}
    q^{n+1} = f(\hat{q}^{n+1}, R^n) \xrightarrow{} a,
\end{equation}
where we ensure that
\begin{equation}
    \hat{q}^{n+1} = f(i_T^{n+1}, R^{n+1})  \xrightarrow{} i_T^n.
\end{equation}
Here, \(i_T^{n+1}\) denotes a new tool input index derived from \(i_T^n\) through reversible operations (e.g., retrieving lyrics from a song name or vice versa). To obtain \(i_T^{n+1}\) and its corresponding relationship \(R^{n+1}\), we employ a search agent that retrieves supersets of \(i_T^n\) to mitigate cyclic generation risks. Specifically, the agent extracts textual content \(C^{n+1}\) as superset candidates, expanding contextual coverage. An LLM then analyzes \(C^{n+1}\) to derive the superset index \(i_T^{n+1}\) and its relationship \(R^{n+1}\) with \(i_T^n\).  
This process ensures progressive context expansion and effective information association. The resulting \(i_T^{n+1}\) and \(R^{n+1}\) are synthesized into an intermediate question candidate \(\hat{q}^{n+1}\), which undergoes rigorous verification. Upon verification, the system generates the refined question \(q^{n+1}\) by integrating \(\hat{q}^{n+1}\) with all historical relationships \(\{R^1,R^2,...,R^n\}\).



\noindent\textbf{Width-based extension.} The goal of the width-based extension is to generate a new task that needs to be decoupled into multiple subtasks to be completed. For simplicity, for two subtasks $q_1 \xrightarrow{} a_1 $ and $q_2 \xrightarrow{} a_2 $, the combined task $q_{width}$ can be represented as 
\begin{equation}
    (q_{width} = q_1+q_2) \xrightarrow{} a_1+a_2,
\end{equation}
where the $+$ indicates using LLM to merge and rephrase two question strings.

\begin{figure*}[htbp]
    \centering
    \includegraphics[width=\textwidth]{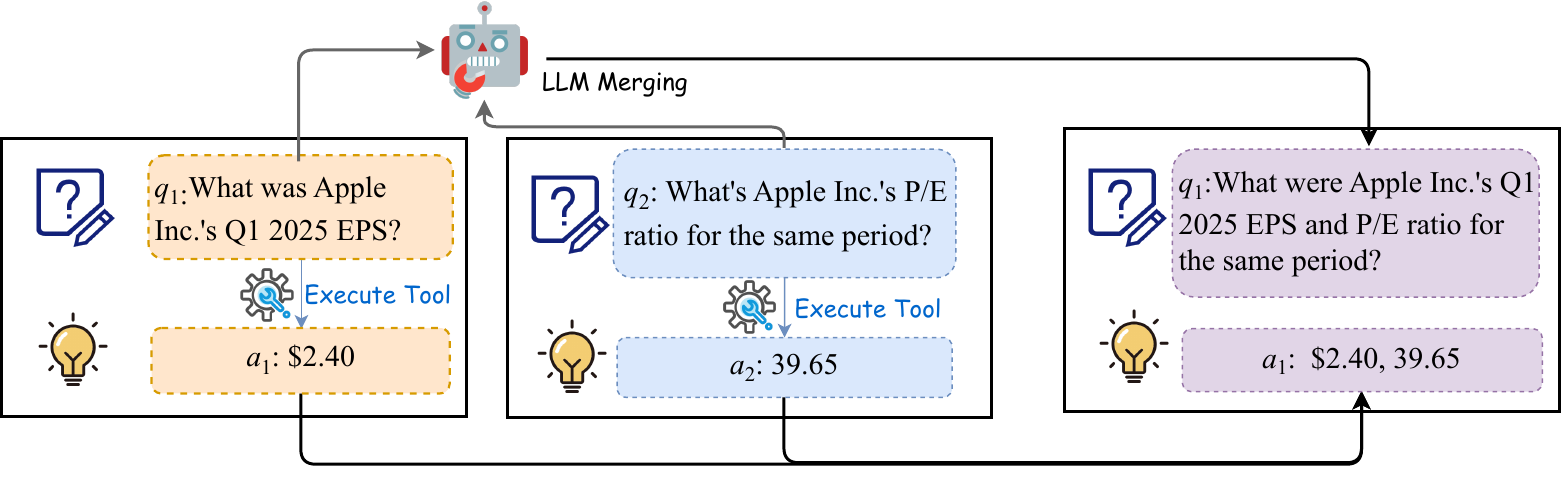}
   \caption{Width-based extension. A new task is formed by merging two subtasks $q_1$ and $q_2$, creating $q_{width} = q_1 + q_2$, where $+$ denotes LLM-based rephrasing.}
   \label{fig:width-merged}
\end{figure*}

\noindent\textbf{Trajectory generation.} Two strategies exist for generating execution trajectories in this task: (1) For simple tasks, such as atomic tasks, existing agents can directly infer and capture the trajectory, including tool selection, parameters, return results, and plans. (2) For complex tasks, such as depth-wise extension tasks, the trajectory is recorded while iteratively expanding and validating new atomic tasks. At each step, the LLM refines the plan or reasoning based on generated intermediate questions.



\subsection{Task Verification} 
Under this generation workflow, the verification of generated tasks can be easily performed in two distinct phases:


\noindent\textbf{Atomic task verification}\label{3.3Atomic task verification}: An atomic task is defined as a simple agent task solvable via a single tool call. During verification, we relax this definition slightly: for each candidate task, we evaluate the task agent’s output within a limited number of tool-use steps (e.g., three) and compare it with an infer-LLM separately. A judge-LLM verifies whether only the agent’s output contains the golden answer, retaining only validated tasks. (see Appendix \ref{appendix:C}  for more details)

\noindent\textbf{Task extension verification}: This process is conducted purely through linguistic analysis without agent involvement. During depth-wise extension, we first employ a judge-LLM to validate: (1) whether the obtained $i_T^{n+1}$ and its relation $R^{n+1}$ constitute a proper superset of $i_T^n$ with logically sound relationships, and (2) whether the final input index $i_T^n$ in $q^n$ is appropriately replaced by $\hat{q}^{n+1}$ in the expanded task $q^{n+1}$. Furthermore, an infer-LLM derives the merged task, while the judge-LLM filters out tasks where the correct result is easily inferred, preventing information leakage that could render the problem trivially solvable after merging.(see Appendix \ref{appendix:B}  for more details). 

This framework ensures efficiency by applying agent reasoning only in atomic task verification at creation, while relying on LLM-based verification elsewhere for faster execution. It also enables complex task generation beyond agent capabilities, with reverse reasoning providing supervisory signals to enhance agent learning or reinforcement learning.

\section{Experiments}
\subsection{Corpus Construction}
    
\begin{figure}[htbp]
    \centering
    \includegraphics[width=0.45\textwidth]{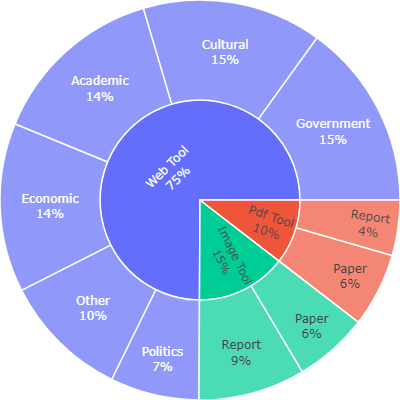}
    \caption{Corpus source distribution. Webpages, PDFs, and images are processed to construct tool-specific tasks.} 
    \label{fig: task data source}
\end{figure}

We collect seed documents across multiple modalities to generate tool-specific atomic tasks, extracting key insights to ensure task relevance. For instance, our PDF processor constructs atomic tasks by combining document titles with core findings, thereby enhancing the necessity for agent-based PDF tool invocation.
To support atomic task generation, we constructed a dataset comprising webpages, PDF files, and images. Webpage data constitutes the largest proportion (75\%), sourced from up-to-date news across multiple domains. Image data accounts for 15\%, primarily derived from financial reports and research papers, with filtering to retain images containing information beyond text. PDF data makes up 10\%, originating from English financial documents and academic publications.

\subsection{Synthetic Tasks Analysis}

\noindent\textbf{}


 
\noindent\textbf{Agent reasoning analysis
.} To practically assess task difficulty, we sample 1,000 tasks and deploy both Smolagents~\cite{smolagents} and its enhanced variant, Smolagents+ (see \secref{appendix:smol} for more details), for execution and validation. While both agents performed identical tasks, Smolagents+ incorporated advanced tool capabilities for refined analysis.

\begin{figure}[htbp]
    \centering
    \includegraphics[width=0.5\textwidth]{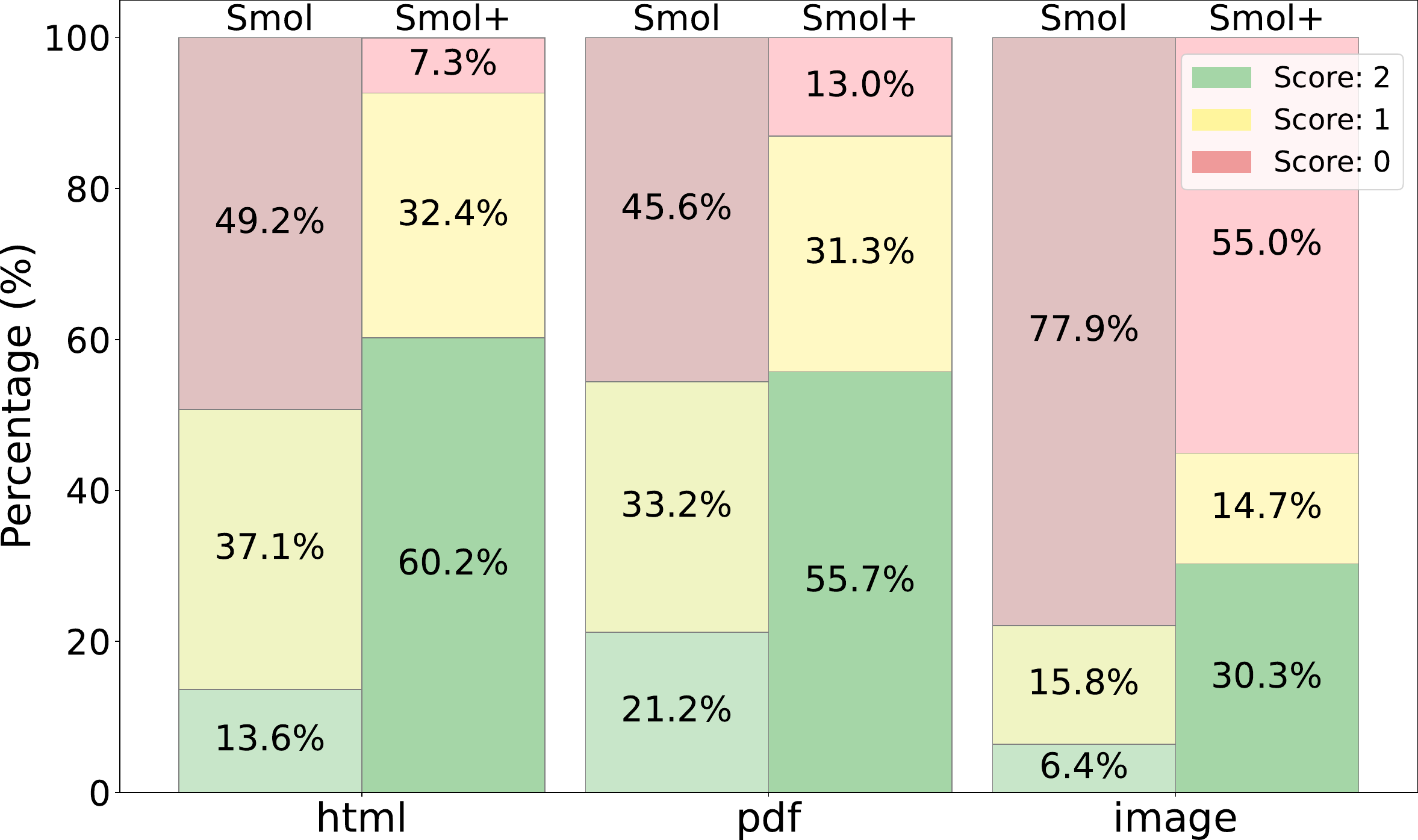}
    \caption{score distribution comparison}  
    \label{fig:agent_score}
\end{figure}

Responses were evaluated by comparing the agents' outputs to the golden answer, following a three-point scoring scheme: 2 for fully correct responses, 1 for answers that included the golden answer but contained additional information, and 0 for incorrect responses.

In \figref{fig:agent_score}, task failure rates increase from web pages to PDFs and then to images within PDFs, indicating that multi-hop web search tasks are more manageable for agents, while complex comprehension challenges, such as PDF extraction and image interpretation, remain difficult.
Additionally, these results demonstrate that our generated tasks span varying difficulty levels, including those that pose significant challenges for current agent capabilities.

\noindent\textbf{Comparison with the GAIA dataset.} \tabref{tab:GAIA} presents the accuracy comparison of Smolagent on the GAIA dataset and our generated dataset. The results indicate that tasks derived from different tool corpora align with GAIA’s varying difficulty levels, with image understanding tasks posing the greatest challenge and achieving accuracy comparable to LEVEL3 data.

\begin{table}[h]
\centering
\caption{Accuracy comparison of Smolagents on the GAIA dataset and our synthetic tasks.}
\resizebox{0.48\textwidth}{!}{
\begin{tabular}{c|cccc}
\toprule
\multirow{2}{*}{GAIA}       & Level1 & Level2 & Level3 & Avg.  \\
 & 54.71  & 43.02  & 26.92  & 44.20 \\
\hline
\multirow{2}{*}{Synthetic Task} & PDF   & html    & Image  & Avg.  \\
 &   54.4  & 50.7   & 22.1   & 42.4 \\
\bottomrule
\end{tabular}
}
\label{tab:GAIA}
\end{table}

Unlike GAIA, which requires extensive human annotation, our approach automates task generation, eliminating the need for labor-intensive data labeling while maintaining scalability and adaptability for agent self-evolution and optimization.

\begin{figure*}[tbp]
    \centering
    \includegraphics[width=1\textwidth]{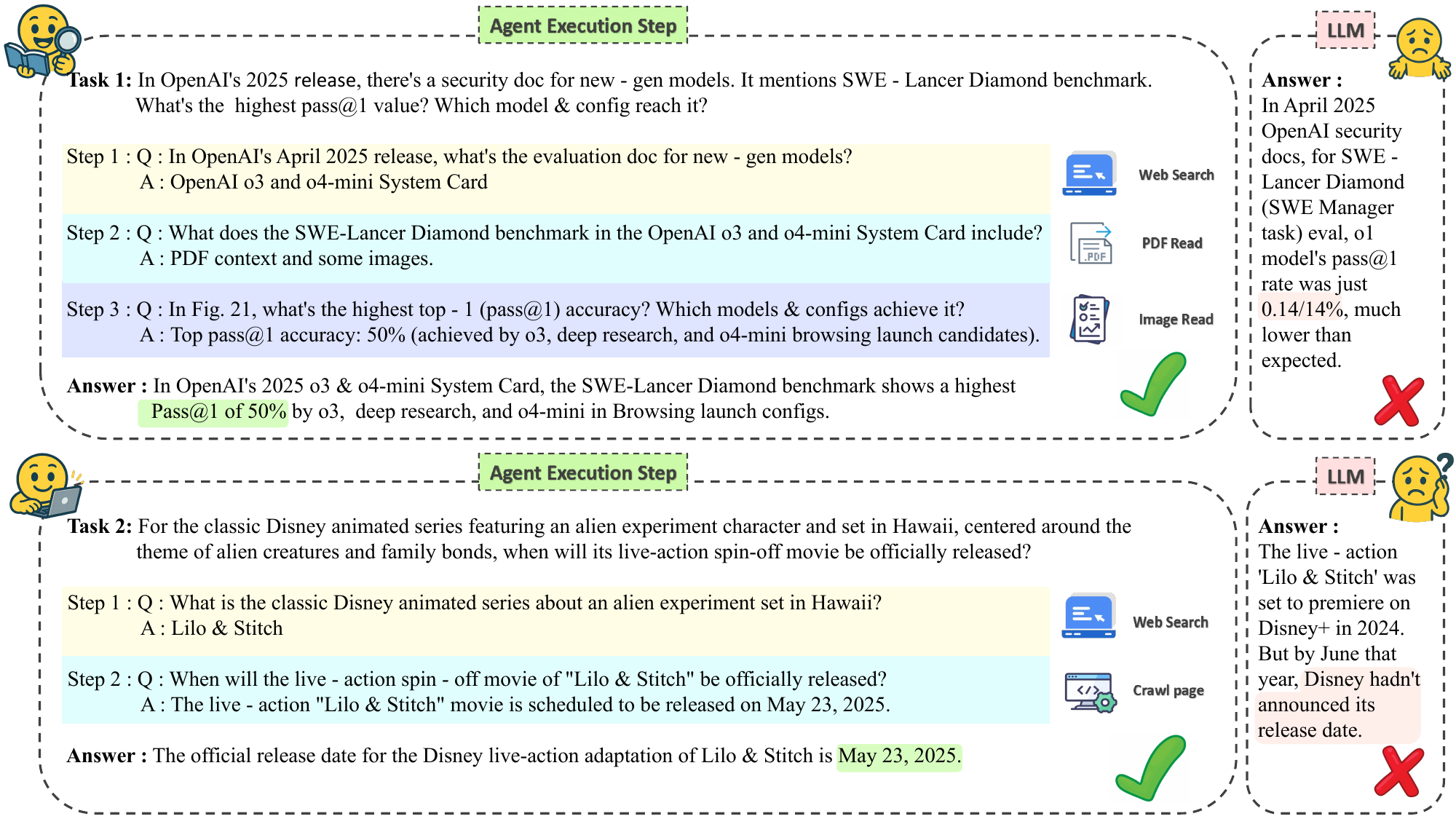}
   \caption{Generated case examples requiring multiple tool calls for completion.}  
   \label{fig:case_study}               
\end{figure*}

\subsection{Enhancing Task Generation Efficiency via Prompt Learning}

We employ rejection sampling in both atomic task generation and task extension. To reduce the rejection rate and enhance sampling efficiency, several key challenges must be addressed:  
\begin{itemize}
    \item Efficiently extract candidate answers from the corpus to support atomic task formation and minimize rejections (\secref{sec:atomic}).
    \item Guide the agent to find an input index $i_T^{n+1}$, ensuring coherent depth-wise extension.
    \item Prompt the LLM in depth-wise extension to articulate the relationship $R^{n+1}$ between the previous input index $i_T^n$ and observed content $C^{n+1}$, refining problem construction and mitigating incoherence-related rejections.
    \item Integrate tasks to ensure precise substitution, i.e., $ q^{n+1} = f(\hat{q}^{n+1}, R^n)$, and clarity while maintaining logical coherence.
\end{itemize}

\noindent\textbf{Evaluation.} We assess atomic task generation and task extension separately.  
For atomic task generation, we evaluate three key metrics: (1) pass rate, representing the proportion of successfully validated atomic tasks relative to candidate tasks. 
(2) task density, quantifying the average number of validated atomic tasks per document. 
(3) sampling time, measuring the time required for processing each document. 

For task extension, we evaluate three key metrics: 
(1) pass rate, the proportion of successful extensions across $n_k$
 attempts (set to 6 in our experiment).
(2) sampling time, measuring the time required for extending each task.  




\noindent\textbf{Prompt Learning.} \label{Prompt Learning}Intuitively, providing the LLM with effective exemplars can further enhance its ability to identify intermediate objectives. To this end, we employ bootstrap few-shot learning~\cite{khattab2024dspy} to systematically optimize the four prompts corresponding to the aforementioned challenges, thereby facilitating the generated workflow. 

For atomic task generation, each prompt is optimized by appending 20 randomly sampled examples. Multiple prompt configurations are then generated by varying these samples, followed by an iterative evaluation process where pass rates determine the optimal selection of inserted examples. 
For task extension, we focus on depth-wise extension and adopt a similar strategy to optimize the prompts using 10 randomly sampled examples. These prompts are refined to maximize the number of hops.

\begin{table}[h]
    \centering
    \caption{Effectiveness of generated task data in prompt learning and depth-wise extension across six extension attempts.}
    \begin{tabular}{lcc}
    \toprule
    Method & Pass rate & Time  \\
    \hline
    Atomic Task & 54.9\% & 29.1s \\
    \textbf{+ Optimization} & \textbf{68.1\%} & \textbf{23.5s} \\
        \hline
    Depth-wise@6 & 41.0\% & 31.5s \\
    \textbf{+ Optimization} & \textbf{51.2\%} & \textbf{30.2s}  \\
    \bottomrule
    \end{tabular}
    \label{tab:agent_opt}
\end{table}

\noindent\textbf{Results.} \tabref{tab:agent_opt} examines atomic task generation and depth-wise task extension before and after prompt learning, highlighting the role of generated task data in enabling self-evolution within both workflows. For atomic task generation, the data improves efficiency by reducing generation time by $19.2\%$ (29.1 to 23.5 seconds) and increasing pass rate from $54.9\%$ to $68.1\%$. Similarly, depth-wise extension benefits from the data, with pass rate rising by $10.2\%$ ($41.0\%$ to $51.2\%$) across six extension attempts, and generation time decreasing by 1.3 seconds (31.5 to 30.2 seconds). These results validate the effectiveness of generated task data in enhancing sampling efficiency and supporting workflow adaptation. The optimized prompts are presented in Appendix \ref{appendix:C optimized prompts}.

\subsection{Fine-Tuning Agent Models Using Synthetic Trajectory}
To validate the effectiveness of our synthetic multi-hop data method, we apply supervised fine-tuning (SFT) and reinforcement learning (RL) using the generated trajectory, refining an agent foundation model—an LLM with tool-integrated reasoning.

\noindent\textbf{Evaluation.} We evaluate our models on three multi-hop question answering benchmark datasets, as follows: HotpotQA \cite{hotpotQA}, Musique \cite{Musique}, and Bamboogle \cite{bamboogle}. These datasets encompass a diverse range of search with reasoning challenges, enabling a comprehensive evaluation.

\noindent\textbf{Baselines.} We conduct a comprehensive evaluation by comparing various baseline models before and after SFT with generated tasks to assess performance improvements: (1) \emph{Base workflow}: We implement agent workflows (Search-R1 without training) across different LLM models. (2) \emph{Search-R1}: An agentic workflow leveraging reinforcement learning for LLM model optimization. 

\noindent\textbf{Implementation setup.} We evaluate two model variants: Qwen2.5-3B-Base and Qwen2.5-3B-Instruct. To facilitate multi-hop reasoning, we synthesize 3,202 multi-hop tasks and their trajectories for SFT. Following the Chain-of-Action framework \cite{coa}, we apply content masking to search tool contexts during training. Our search method, RL training data, and reinforcement learning strategy follow the Search-R1 \cite{search-r1}. For further training details, refer to Appendix \ref{More_Implementation_Details}.

\begin{table}[h]
\centering
\resizebox{0.495\textwidth}{!}{
\begin{tabular}{lcccc}
\hline
Method           & HotpotQA & Musique & Bamboogle & Avg. \\ \hline
\multicolumn{4}{l}{\textbf{Qwen2.5-3b-Base}}      \\
Base workflow & 0.032    & 0.006   & 0.063  & 0.034   \\
\textbf{+ SFT} & \textbf{0.232}    & \textbf{0.067}   & \textbf{0.224}  & \textbf{0.174}   \\
Search-R1& 0.284    & 0.049   & 0.088  &  0.140  \\
\textbf{+ SFT}  & \textbf{0.344}    &\textbf{ 0.111}   &\textbf{ 0.280 }  & \textbf{0.245}  \\ \hline
\multicolumn{4}{l}{\textbf{Qwen2.5-3b-Instruct}}  \\
Base workflow & 0.190    & 0.037   & 0.112  & 0.113   \\
\textbf{+ SFT} &\textbf{ 0.221}    &\textbf{ 0.049}   & \textbf{0.248 } & \textbf{0.173}   \\
Search-R1         & 0.324    & 0.103   & 0.264  & 0.230   \\
\textbf{+ SFT}  & \textbf{0.340}    &\textbf{ 0.104}   & \textbf{0.264}  & \textbf{0.236}   \\ \hline
\end{tabular}
}
\caption{Performance across three datasets and two models. Avg. denotes average.}
\label{tab:sft}
\end{table}
 
\noindent\textbf{Results.} As shown in \tabref{tab:sft}, our method demonstrates significant performance improvements across three representative datasets and two model variants. 

First, our synthetic data demonstrates significant value in standalone SFT training, achieving average performance improvements of +14.0\% (Qwen2.5-3B-Base) and +6.0\% (Qwen2.5-3B-Instruct) over the base workflow for their respective models. These gains validate the quality and effectiveness of our synthetic data generation methodology.

Second, compared to the Search-R1 baseline, the workflow with Qwen2.5-3b-Base achieves maximum gains of +19.2\% on Bamboogle and +6.2\% on Musique. The Qwen2.5-3B-Instruct maintains steady gains, with an average performance margin of +0.6\%.
The strong performance of our SFT-trained models underscores their suitability for subsequent reinforcement learning, suggesting that our synthetic data not only enhances immediate task execution but also provides a more effective initialization for RL optimization.

\subsection{Effectiveness of Tool Context in Constructing Agentic Tasks.}
\label{sec:abla}
In atomic task generation, we integrate the additional input index $i_T$ along with the relational mapping $R$ between the tool context and a given answer to systematically structure tasks.

To assess the efficiency of our atomic task generation approach, we perform an ablation study using an LLM to directly generate a task $q$ that requires only one external tool to obtain the answer $a$, explicitly excluding the conditions $i_T$ and $R$. Evaluation metrics include pass rate, task resolution time, average tool usage, and the variance in tool usage frequency.

\begin{table}[h]
    \centering
    \caption{The effectiveness of tool context.}
    \resizebox{0.48\textwidth}{!}{
    \begin{tabular}{lcccc}
    \toprule
    Method & Pass rate & Time & \#Tool-use & $\sigma^2$\\
    \hline
    LLM only & 18.5\% & 119.7s & 2.8 & 1.2 \\
    \textbf{Ours} & \textbf{43.0\%} & \textbf{86.7s} & \textbf{2.1} & \textbf{0.4}\\
    \bottomrule
    \end{tabular}
    }
    \label{tab:abla}
\end{table}

Compared to atomic tasks generated via direct prompting of GPT-4.1, our approach significantly enhances atomic task generation efficiency. Specifically, our workflow achieves a 24.5\% higher pass rate (43.0\% vs. 18.5\%) while reducing task generation time by 28 seconds (86.7s vs. 119.7s), underscoring the limitations of vanilla LLMs in constructing agentic tasks. Furthermore, our atomic tasks exhibit greater atomicity, as evidenced by a lower average tool invocation count (2.1 vs. 2.8 per query). Task complexity also remains more stable and controllable, with a reduced variance in tool usage (0.4 vs. 1.2). These findings underscore the robustness of our workflow, validating its efficacy in structured task generation.

\section{Related Work}
\subsection{Instruction Data Generation}
Synthetic data has emerged as a promising solution for enhancing performance and enabling new capabilities. STaR~\cite{zelikman2024star} augments learning with chain-of-thought (CoT) rationales but often requires a substantial number of task queries beforehand. Methods such as Self-Instruct~\cite{wang2022self}, Self-Chat~\cite{xu2023baize}, NuminaMath~\cite{li2024numinamath}, and OpenMathInstruct-2~\cite{toshniwal2024openmathinstruct} generate data from minimal seed examples using LLMs, yet they struggle to extend task generation for multiple tool invocations.
WizardLM~\cite{xu2023wizardlm} employs Evol-Instruct to incrementally enhance instruction complexity. However, it relies primarily on rule-based modifications, making its generated instructions unsuitable for agentic task scenarios.
MetaMath~\cite{yu2023metamath} generates mathematical data by rewriting questions, but adapting agent tasks to environmental feedback presents challenges beyond simple rephrasing. WebInstruct~\cite{yue2024mammoth2} extracts question-answer pairs from a pre-training corpus across multiple domains; however, the generated questions often fail to incorporate tool utilization in their solutions. AutoAct~\cite{qiao-etal-2024-autoact} uses a self-planning mechanism to generate planning trajectories for QA tasks.

\subsection{Language Agent}
Existing research on agentic task execution primarily advances along two core dimensions: role specialization and functional partitioning. Role-based paradigms structure collaborative networks by dynamically allocating differentiated tools, as demonstrated by AutoGPT~\cite{autogpt2023autogpt}, AutoGen~\cite{wu2023autogen}, and Camel~\citep{li2023camel}. In contrast, functional partitioning frameworks, such as Barcelona2, Omne, and AgentIM~\footnote{These are closed-source frameworks.}, define distinct task execution roles, optimizing modular efficiency.
Smolagents~\citep{smolagents} combines the ReAct~\citep{yao2023react} and CodeAct~\citep{wang2024executable} architectures to build a multi-functional agents hierarchy to perform multiple rounds of interactions and actions in code to accomplish complex tasks. Magnetic-One~\cite{fourney2024magentic} refines vision-language processing by decoupling perception~\cite{yang2023how2comm,yang2023what2comm}, planning~\cite{song2023llm,tordesillas2021mader}, and execution modules~\cite{qin2024tool,wang2024executable}, improving efficiency in multimodal environments.
Dynamic orchestration mechanisms address real-time task reallocation and system resilience. Trase-Agent~\cite{trase2024trase} adapts execution strategies based on real-time feedback, while TapeAgents~\cite{bahdanau2024tapeagentsholisticframeworkagent} employs asynchronous communication to enhance robustness in agent coordination. Empirical findings suggest that stabilized sub-agent interactions yield higher task success rates than complex, centralized orchestration algorithms.

To further extend agentic autonomy, AutoAgent~\cite{tang2025autoagent} facilitates intelligent execution and personalized agent customization without requiring manual coding. Its core components—natural language-driven coordination, customizable workflows, and self-managing file systems—streamline agent development. Hybrid architectures, such as h2oGPTe-Agent~\cite{h2oGPTe2024h2oGPTe}, explore multi-agent optimization strategies, achieving over 70\% accuracy in code generation tasks. However, significant cross-modal processing bottlenecks remain an open challenge.
\section{Conclusion}
We present \textsc{TaskCraft}, an automated workflow for scalable, multi-tool, verifiable agentic task generation. Through width-based and depth-based extension, our framework constructs hierarchically complex challenges. Empirical results demonstrate its effectiveness in structured task generation, improving prompt optimization and supervised fine-tuning while reducing reliance on human annotation. Additionally, we release a large-scale synthetic dataset of approximately 36,000 tasks with varying difficulty to support future research on agent tuning and evaluation.

\section{Limitation}
This work currently focuses on constructing atomic tasks for common tools, including browsing, PDF processing, and image analysis. Future iterations will enable users to generate atomic tasks tailored to their agents' specific tool requirements. 

\clearpage
\section*{Contributions}

\textbf{Core Contributors}
\begin{multicols}{2}
\begin{itemize}
    \item Dingfeng Shi
    \item Qianben Chen

    \item Jingyi Cao
\end{itemize}
\end{multicols}

\textbf{Contributors}

\begin{multicols}{2}
\begin{itemize}
    \item Weichen Sun
    \item Hongxuan Lu
    \item Tianrui Qin
    \item Minghao Liu
    \item Ge Zhang
    \item Changwang Zhang
    \item Yuchen Eleanor Jiang

    \item Weizhen Li
    \item Fangchen Dong
    \item King Zhu
    \item Jian Yang
    \item Jiaheng Liu
    \item Jun Wang
    \item[]
\end{itemize}
\end{multicols}

\textbf{Corresponding Authors}
\begin{multicols}{1}
\begin{itemize}
\item Wangchunshu Zhou
\end{itemize}
\end{multicols}
\clearpage
\bibliographystyle{plainnat}
\bibliography{ref.bib}

\clearpage

\beginappendix

\appendix

\section{Data Statistics}

\begin{figure}[htbp]
    \centering
    \includegraphics[width=0.5\textwidth]{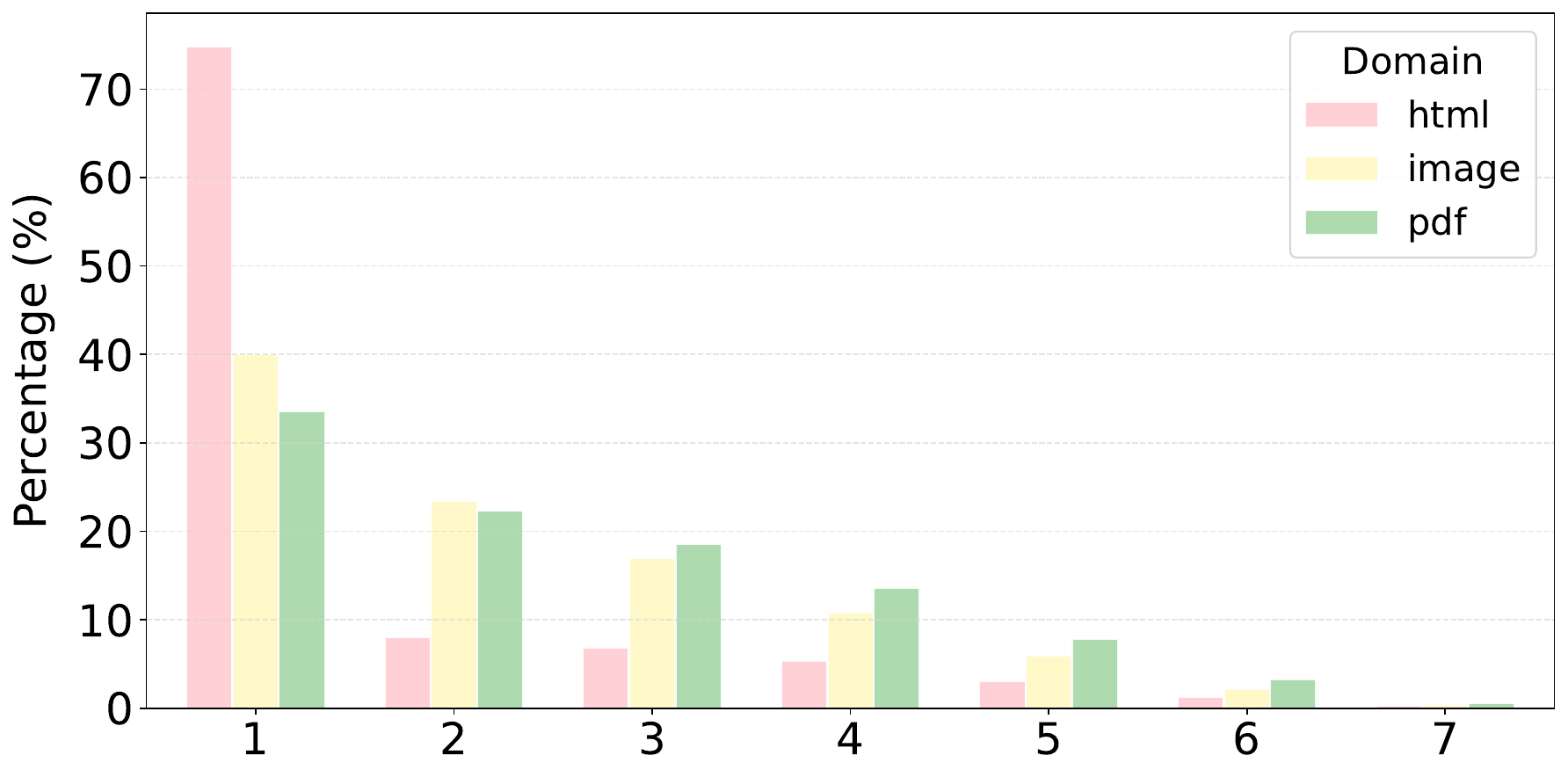}
   \caption{Analysis of all tasks.}
   \label{fig:data_ana}
\end{figure}

As illustrated in \figref{fig:data_ana}, task generation exhibits a hierarchical decay pattern across all domains as hop count increases, revealing distinct scalability trends:

\begin{itemize}
\item \textbf{PDF domain}: Shows gradual performance attenuation with hop depth, with 1-hop tasks accounting for 33.62\% (2,737 tasks), decreasing to 22.36\% (1,820 tasks) for 2-hop and 18.60\% (1,514 tasks) for 3-hop. The sharp drop in 5-7 hop tasks (11.80\% combined) indicates limited deep-extension capability, yet still surpasses other domains in depth scalability.

\item \textbf{Image domain}: Presents the most pronounced performance decay, with 1-3 hops comprising 80.45\% (4,342/5,397 tasks) but only 8.64\% (467 tasks) for 5-7 hops, highlighting fundamental constraints in deep hierarchical task generation.

\item \textbf{HTML domain}: In the HTML domain, 1-hop tasks dominate, constituting 74.84\% (17,154 tasks) of the total. However, this domain also has the highest absolute number of deep extensions, with 5-7 hop tasks accounting for 4.75\% (1,089 tasks).
\end{itemize}


\begin{figure}[htbp]
    \centering
    \includegraphics[width=0.37\textwidth]{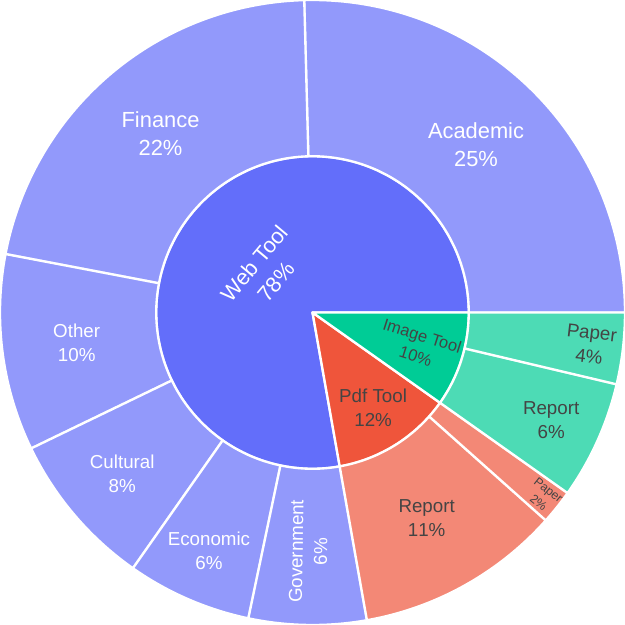}
    \caption{Distribution of atomic data.}  
    \label{fig: task data source}
\end{figure}
\noindent\textbf{Atomic task analysis.} We collect data from webpages, PDF files, and images to support the generation of atomic tasks, which form the basis of the dataset, totaling 22,053 instances as shown in Figure 8.

Among them, atomic conclusions extracted by web-based tools account for the largest proportion, reaching 77.78\%, with sources spanning multiple domains: academic (25.42\%), financial (21.58\%), cultural (8.09\%), economic (6.45\%), and governmental (6.08\%) resources. These conclusions are derived from up-to-date news and curated online materials to ensure relevance.

Image-based tools contribute 9.80\% of the data, primarily extracting structured insights (e.g., key trends, comparisons) from charts and tables in financial reports and research papers. To avoid redundancy, we implement strict verification to exclude conclusions that directly replicate source text.

PDF-based extraction accounts for 12.41\%, supplementing the dataset with findings from financial reports and academic publications. This multi-source approach enhances diversity while maintaining consistency in atomic fact representation.

By systematically integrating these extraction methods, we ensure high-quality task generation, providing a robust foundation for downstream model training and optimization.

\section{Verification Requirements for Depth-Based Extension}\label{appendix:B}

Effective n-hop task extension requires rigorous verification to ensure valid multi-hop reasoning. The transformation must preserve superset validity:

\begin{equation}
\hat{q}^{n+1} = f(i_T^{n+1}, R^{n+1}) \rightarrow i_T^n
\end{equation}

\begin{equation}
    q^{n+1} = f(\hat{q}^{n+1}, R^n) \xrightarrow{} a
\end{equation}

Current depth-based extension methods often introduce two critical flaws when replacing tool inputs $i_T$ without proper verification:

\begin{itemize}
\item \textbf{Pseudo-Superset Problem}: Superficial substitutions that preserve semantic equivalence but lack genuine superset relationships
\item \textbf{Information Leakage}: Premature disclosure of information that should only emerge through proper multi-step reasoning
\end{itemize}
 These issues undermine the intended multi-hop reasoning process.

\subsection{Pseudo-Superset Problem}

A fundamental limitation arises when replacing $i_T$ with a semantically equivalent but non-superset index $i_T^{n+1}$. Consider the following task extension example:  
\begin{AIbox}
    {\makecell{Original task}}{
    \textbf{Query ($q^n$):}
How many travel trends for 2022 does 'Travel Trends 2025 | Our Annual Report' present?

\textbf{Answer: } 5
}
\end{AIbox}




Substituting $i_T$ ( "Travel Trends 2025 | Our Annual Report") with the synonymous $i_T^{n+1}$ ("2025 Annual Travel Trends Report") yields a intermediate task:

\begin{AIbox}
    {\makecell{Intermediate task}}{
\textbf{Query ($\hat{q}^{n+1}$):}
What is the title of 2025 Annual Travel Trends Report?

\textbf{Answer :} Travel Trends 2025
}
\end{AIbox}
Despite valid hop annotations, the intermediate question does not constitute an effective extension: it does not represent a necessary tool-use step. The core issue lies in the absence of a genuine superset relationship between \(i_T^n\) and \(i_T^{n+1}\), leading to superficial expansion.
\begin{AIbox}
    {\makecell{Extended task}}{
\textbf{Query ($q^{n+1}$): }
How many travel trends for 2022 does '2025 Annual Travel Trends Report' present?

\textbf{Answer: } 5
}
\end{AIbox}
\subsection{Information Leakage}

A second failure mode occurs when expanded tasks inadvertently expose original answers, enabling large language models (LLMs) to bypass tool retrieval. For instance, consider the extended task:

\begin{AIbox}
    {\makecell{Extended task}}{
\textbf{Query ($q^{n+1}$):} In the AP Sports daily summary, Charter and Cox's proposed merger is valued at approximately \$34.5 billion. What is the exact amount?

\textbf{Answer :} 34.5B USD
}
\end{AIbox}
While this query appropriately conceals the previous $i_T^n$ ("Sports In Brief"), it directly reveals the answer "34.5B USD", allowing the LLM to bypass the intended retrieval process. This compromises the essential tool dependency required for multi-hop task answering.

\subsection{Verification for Task Extension}

To address these challenges, we propose a rigorous verification framework to ensure the validity of $i_T^{n+1}$, $\hat{q}^{n+1}$ and $q^{n+1}$ in task extension.

\subsubsection{Strict Superset Verification}
\label{appendix B:Superset}

$i_T^{n+1}$ must be the index of a strict superset of $i_T^n$, and the relationship can be formalized as:

\begin{equation}
\hat{q}^{n+1} = f(i_T^{n+1}, R^{n+1}) \rightarrow i_T^n
\end{equation}

where $R^{n+1}$ denotes hierarchical relations (e.g., \textit{contains}, \textit{part\_of}). Valid extensions must introduce genuine depth, such as \textit{"Sports In Brief"} $\rightarrow$ \textit{"AP News's Sports Section"} (relation: \textit{contains}), while rejecting synonymous substitutions. Additionally, invalid extensions that allow the LLM to derive $i_T^{n}$ directly should be excluded.

\subsubsection{Information Leakage Verification}
\begin{equation}
    q^{n+1} = f(\hat{q}^{n+1}, R^n) \xrightarrow{} a
\end{equation}

The extended query $q^{n+1}$ must adhere to the information-sealing principle to ensure proper tool-use reasoning. This requires that the query does not directly expose the original answer, and any query from which the LLM can directly obtain the answer should be filtered out. 

\subsection{Advantages of the Verification Framework}

Our approach provides three key advantages:
\begin{itemize}
\item \textbf{Superset Integrity}: Guarantees valid hierarchical progression (e.g., \textit{column} $\rightarrow$ \textit{page} $\rightarrow$ \textit{website}) without logical gaps.
\item \textbf{Strict Tool Dependency}: Enforces authentic multi-hop reasoning by eliminating solution shortcuts, ensuring mandatory tool-use.
\item \textbf{Transparent Reasoning}: Offers full explainability through explicit relation paths ($R^n$).
\end{itemize}

A properly expanded task under this framework would appear as follows:
\begin{AIbox}
    {\makecell{Qualified Extended task}}{
\textbf{Query ($q^{n+1}$):}
According to the recurring AP News's sports section feature that regularly provides concise summaries of top sports events and highlights, what is the merger value currently being pursued by US cable giants Charter and Cox as they face increasing competition from streaming services?

\textbf{Answer :} 34.5B USD
}
\end{AIbox}

\section{Core Prompts}\label{appendix:C}
This section presents key components of the verification prompts used in our framework.

\subsection{Atomic task verification}
The following prompt is used in atomic task verification (Section~\ref{3.3Atomic task verification}):

\begin{AIbox}{Atomic task verification}
\textbf{Task}: Evaluate the \textit{consistency} between the golden answer (GA) and another answer (AA, either agent or LLM-generated) as follows:

\begin{itemize}
    \item \textbf{2 points (Fully Consistent)}: AA and GA are semantically equivalent, even if phrased differently. \\
    \textit{Example}: 
    \begin{itemize}
        \item GA: ``Interest rates should be raised and inflation monitored.''
        \item AA: ``It is necessary to raise interest rates and monitor inflation.''
    \end{itemize}
    
    \item \textbf{1 point (Partially Consistent)}: AA includes all GA information but adds valid extra details. \\
    \textit{Example}: 
    \begin{itemize}
        \item GA: ``The interest rates should be raised.''
        \item AA: ``The interest rates should be raised, and inflation monitored.''
    \end{itemize}
    
    \item \textbf{0 points (Inconsistent)}: AA omits key GA information or contradicts it. \\
    \textit{Examples}: 
    \begin{itemize}
        \item \textit{Omission}: GA: ``Raise rates and monitor inflation.'' \\ AA: ``Raise rates.''
        \item \textit{Contradiction}: GA: ``Raise rates by 50bps.'' \\ AA: ``Raise rates by 25bps.''
    \end{itemize}
\end{itemize}

\noindent The criteria prioritize semantic equivalence while accommodating informative expansions or reductions.

\textbf{Output Format}: ...
\end{AIbox}
A task is retained as an atomic task if and only if: (1) the \textit{AgentScore} strictly exceeds the \textit{LLMScore}, and (2) the \textit{AgentAnswer} is non-zero.
\subsection{optimized prompts}
\label{appendix:C optimized prompts}
The following prompts is optimized prompt mentioned in (Section~\ref{Prompt Learning}):

\begin{AIbox}{Atomic Conclusion Extraction}
\textbf{Task}: Extract standalone conclusions from document chunks meeting these criteria:

    
    
    
    
    

\begin{enumerate}
    \item \textbf{Atomicity}: Extract only indivisible basic facts (no combined conclusions, e.g., split ``A increased by 5\% and B decreased by 2\%'' into two separate conclusions)
    
    \item \textbf{Verifiability}: Include at least one definite identifier (numeric value, time, unique name) and reject vague expressions (e.g., ``Performance has improved'')
    
    \item \textbf{Timeliness Handling}: Explicitly mark time ranges for time-sensitive information (e.g., ``Global GDP grew by 3.0\% in 2023'' instead of ``Recent GDP growth of 3.0\%'')
    
    \item \textbf{Citation Integrity}: Embed complete content of cited references (e.g., expand ``as stated in (2)'' to include the full text of (2) in the conclusion)
\end{enumerate}

\textbf{Valid Examples}:
\begin{itemize}
    \item \textbf{Example 1:} 3D deconvolution microscopy illumination optimization for refractive index tomography (Optics Express 29, 6293-6301, 2021)
    \item \textbf{Example 2:} Azimuthal energy $\Phi$ parameters: ($\theta_0=0.5$, $\theta_d=2\pi/7$, $\theta_w=\pi/9$, $\theta_f=0.06$, $p=1.0004$, $q=100$)
    
    \ldots (more examples omitted) \ldots
    
\end{itemize}
\textbf{Output Format}: ...
\end{AIbox}
\begin{AIbox}{Depth-wise Extension: Index $i_T^{n+1}$ Guidance and $R^{n+1}$ Articulation}
\textbf{Task}: Identify a minimal unique superset for an input element based on its attributes, ensuring the superset+relationship uniquely points to the element.

\textbf{Examples}:
\begin{enumerate}
    \item Paragraph/sentence: Its belonging text content
    \item Specific term: Corresponding discipline/category
    \item Specific date: Date range it's in (e.g., its week/month)
    \item Short event: Complete specific event it's part of
    \item Page: Referencing pages or parent page
    \item Generate only one relationship, avoiding strongly specific proper nouns
\end{enumerate}

\textbf{Relationship expression guidelines}:
\begin{enumerate}
    \item Clearly show hierarchical/ownership. Indicate position for series sub-items; clarify ownership for parts of a superset
    \item Specify input content's positioning (e.g., time range, publication field, role in superset)
    \item Use research/industry standard wording
    \item Provide only necessary associations
\end{enumerate}

    

\textbf{Notes}:
\begin{enumerate}
    \item Return the superset’s unique identifier (e.g., attribute name, page title, paper title)
    \item Obtain superset content via tool (web, PDF, image)
    \item Concisely describe the relationship, listing unique qualification conditions
    \item Use $\leq$3 search keywords per search; do multiple searches if needed
    \item Derive the identifier from search results, excluding the input content
    \item Prioritize reading PDF content with tools if the input is a PDF

\end{enumerate}

\textbf{Valid Examples}:
\begin{itemize}
    \item \textbf{Example 1:}
    \begin{itemize}
        \item \textbf{Input:} Avatar 3: Fire and Ash
        \item \textbf{Superset Identifier:} Avatar film series
        \item \textbf{Relation:} The third film
    \end{itemize}

    \item \textbf{Example 2:}
    \begin{itemize}
        \item \textbf{Input:} V3LMA: Visual 3D-enhanced Language Model for Autonomous Driving
        \item \textbf{Superset Index:} cs.CV
        \item \textbf{Relation:} A paper on visual 3D-enhanced language models for autonomous driving
    \end{itemize}
    
    \ldots (more examples omitted)\ldots
\end{itemize}
\textbf{Output Format}: ...
\end{AIbox}
\begin{AIbox}{Logical Substitution: $q^{n+1}$ as $f(\hat{q}^{n+1}{,} R^n)$}
\textbf{Task}: Substitute elements in core queries using auxiliary queries while preserving:


\begin{enumerate}
    \item \textbf{Complexity Balance}: The new query should be slightly more complex than the original core Query and require more steps to solve. But do not make too many changes to the core query.
    
    \item \textbf{Answer Uniqueness}: The new query should point to the unique answer: golden answer, and should not point to other answers.

    \item \textbf{Answer Concealment}: The new query must not reveal information about the golden answer.

    \item \textbf{Natural Language Polish}: After merging, polish the question to make it conform to human expression habits without changing the original meaning. Do not modify the proper nouns appearing in it.
  
\end{enumerate}
\textbf{Valid Examples (20 in total)}:
\begin{itemize}
    \item \textbf{Example 1:}
    \begin{itemize}
        \item \textbf{Core Query:} What is the 2nd positive integer?
        \item \textbf{Auxiliary Query:} Numbers except 0 in natural numbers
        \item \textbf{New Query:} What is the 2nd natural number except 0?
    \end{itemize}

    \item \textbf{Example 2:}
    \begin{itemize}
        \item \textbf{Core Query:} Ne Zha 2 attendance ranking
        \item \textbf{Auxiliary Query:} 2025 May Day box office summary
        \item \textbf{New Query:} Given 2025 May Day box office data, what is Ne Zha 2's attendance ranking?
    \end{itemize}




    \ldots \quad (18 more examples omitted)
\end{itemize}
\textbf{Output Format}: ...
\end{AIbox}
\subsection{Strict Superset Verification}
The following prompt is used in Appendix~\ref{appendix B:Superset}:
\begin{AIbox}{Strict Superset Verification}
\textbf{Task}: Verify if index $i_T^{n+1}$   uniquely determines subset $i_T^n$ under relation $R^n$ in given queries.

\textbf{Criteria}:
\begin{enumerate}
    \item \textbf{Superset\-Subset Relationship}:
    \begin{itemize}
        \item $i_T^{n+1}$ must be the index of a superset that properly contains $i_T^n$
        \item $i_T^{n+1} \not\approx i_T^n$ (excluding synonym pairs like \textsc{Car}/\textsc{Automobile})
    \end{itemize}
    
    
   \item \textbf{Relationship Validity}:
    \begin{itemize}
        \item The relationship $R^n$ must explicitly and uniquely link the superset to the subset (no many-to-one mappings)
    \end{itemize}
    
\end{enumerate}

\textbf{Output Format}: ...
\end{AIbox}

\section{Further Training Detail}
\label{More_Implementation_Details}
For SFT training, we synthesize 3,202 multi-hop tasks and their trajectories and apply content masking to search tool contexts in these trajectories. 

For RL training, we follow the Search-R1 \cite{search-r1} and use the 2018 Wikipedia dump as a knowledge source and the E5 embedding model as a retriever. For fair evaluation, we fix the retrieval depth to 3 passages for all methods. We merge the training sets of NQ and HotpotQA to form a unified dataset. Evaluation is conducted on the test or validation sets of three datasets to assess both in-domain and out-of-domain performance. Exact Match is used as the evaluation metric. In the PPO settings, we set the learning rate of the policy LLM to 1e-6 and that of the value LLM to 1e-5. Training is conducted for 500 steps, with warm-up ratios of 0.285 and 0.015 for the policy and value models, respectively. We use Generalized Advantage Estimation with parameters $\lambda$ = 1 and $\gamma$ = 1. We employ vLLM for efficient LLM rollouts, configured with a tensor parallelism degree of 1 and a GPU memory allocation ratio of 0.6. Our sampling strategy utilizes a temperature parameter of 1.0 and top-p threshold of 1.0. For policy optimization, we apply KL divergence regularization with coefficient $\pi$=0.001 and implement a clip ratio $\epsilon$=0.2. The action budget is constrained to 4, with a default retrieval depth of 3 passages per query.

\section{Smolagents+}
\label{appendix:smol}
We developed Smolagents+, enhancing its web search capabilities, integrating multiple information sources, streamlining search results, and implementing a query rewriting strategy to optimize search performance.

\end{document}